\documentclass[lettersize,journal]{IEEEtran}
\usepackage{amsmath,amsfonts}
\usepackage{array}
\usepackage[caption=false,font=normalsize,labelfont=sf,textfont=sf]{subfig}
\usepackage{textcomp}
\usepackage{stfloats}
\usepackage{url}
\usepackage{verbatim}
\usepackage{graphicx}
\usepackage{cite}
\usepackage{enumerate}
\usepackage{multirow}
\usepackage[misc,geometry]{ifsym}
\usepackage[ruled,linesnumbered]{algorithm2e}
\begin{document}

\title{TS-ENAS:Two-Stage Evolution for Cell-based Network Architecture Search}

\author{Juan~Zou \textsuperscript{1},
	\Letter Shenghong~Wu \textsuperscript{1},
	\Letter Yizhang~Xia,
	Weiwei~Jiang,
    Zeping~Wu,
	and~Jinhua~Zheng
	\thanks{1 These authors have contributed equally to this work.}
	\thanks{Juan Zou, Shenghong Wu (Corresponding author, email: 202021002368@smail.xtu.edu.cn), Yizhang Xia (Corresponding author, email: yizhang.xia@xtu.edu.cn), Weiwei Jiang and Jinhua Zheng are with Engineering Research Center of Hunan Province for Optimization and Security of Intelligent System, Key Laboratory of Intelligent Computing and Information Processing, Ministry of Education of China, and Key Laboratory of Hunan Province for Internet of Things and Information Security, Xiangtan University, Xiangtan, 411105, Hunan Province, China.}
	\thanks{Zeping Wu are with the College of Aerospace Science and Engineering, National University of Defense Technology, Changsha, Hunan, P. R. China, 410073}}


\maketitle

\begin{abstract}
Neural network architecture search provides a solution to the automatic design of network structures. However, it is difficult to search the whole network architecture directly. Although using stacked cells to search neural network architectures is an effective way to reduce the complexity of searching, these methods do not able find the global optimal neural network structure since the number of layers, cells and connection methods is fixed. In this paper, we propose a Two-Stage Evolution for cell-based Network Architecture Search(TS-ENAS), including one-stage searching based on stacked cells and second-stage adjusting these cells. In our algorithm, a new cell-based search space and an effective two-stage encoding method are designed to represent cells and neural network structures. In addition, a cell-based weight inheritance strategy is designed to initialize the weight of the network, which significantly reduces the running time of the algorithm. The proposed methods are extensively tested and compared on four image classification dataset, Fashion-MNIST, CIFAR10, CIFAR100 and ImageNet and compared with 22 state-of-the-art algorithms including hand-designed networks and NAS networks. The experimental results show that TS-ENAS can more effectively find the neural network architecture with comparative performance.
\end{abstract}


\section{Introduction}
\IEEEPARstart{D}{eep} neural networks (DNNs)\cite{res1}, the fundamental component of deep learning, have demonstrated their dependability in a variety of practical tasks, including image classification\cite{ref2}, natural language processing\cite{ref3}, speech recognition\cite{ref4}. Typically, a DNN's effectiveness is influenced by its architecture and optimal weights. In order to reduce the loss of the difference between the output that really occurs and the output that is intended, the optimal weight often employs a gradient-based reduced method. Furthermore, achieving the ideal architecture cannot be described by a continuous function. Therefore, promising DNN architectures are hand-designed with extensive expertise, such as VGG\cite{ref5}, ResNet\cite{ref6}, DenseNet\cite{ref7} and so on. Researchers often spend a lot of time designing a suitable network structure for real-world problems. By considering the construction of DNN structures for target tasks to be an optimization issue, Neural Architecture Search (NAS) attempts to automatically produce appropriate neural network designs for target tasks, offering an effective solution to time-consuming network architecture design activities.

According to theory, NAS is a challenging optimization problem\cite{ref8} with several obstacles, including complicated restrictions, discrete representation, a two-layer structure, computationally expensive features, and numerous competing criteria. An optimization method specifically created to handle such challenging optimization situations is the NAS algorithm. Generally speaking, NAS algorithms can be classified into three groups depending on distinct optimization techniques: NAS algorithms based on reinforcement learning (RL)\cite{ref9}, NAS algorithms based on gradients and NAS algorithms based on evolutionary computing (ENAS)\cite{ref10}. The computationally resource-intensive RL-based NAS approach necessitates training multiple models on a specific dataset for each optimization. For instance, on the CIFAR-10 image classification benchmark dataset\cite{ref11}, the RL-based NAS algorithm\cite{ref12} needs to use 800 graph processing units and spend 28 days to find the best model. Although the gradient-based approach is more successful compared to the RL-based algorithm, this hasn't been formally demonstrated. The gradient-based technique also needs to construct a super network and has relatively high professional knowledge requirements. The ENAS algorithm uses evolutionary computing (EC) technology to solve NAS problems. A tough optimization problem can be solved using EC technology, a group-based evolution paradigm, by imitating the evolution of organisms or the behavior of groups in nature. However, the computational load of the ENAS algorithm is relatively high. For example, on the CIFAR-10 benchmark dataset, the ENAS method\cite{ref13} requires 450 GPUs and takes 7 days to find a deep neural network with a classification accuracy of 96.88\%.

The biggest drawback of NAS is the amount of computing resources is needed to train the neural network. Therefore, more and more researchers have begun to use performance prediction\cite{ref14},\cite{ref15}, hierarchical representation\cite{ref13} and weight sharing\cite{ref16},\cite{ref17} to increase the algorithm's computational efficiency. Based on the model structure, the performance prediction may directly estimate the precision of deep neural network models. An LSTM model was employed by Chenxi Liu et al.\cite{ref18} to direct their search for the network structure.The verification accuracy of the forecast is the output of the model, and its input is a variable-length string description of the network structure. Another strategy is to forecast the neural network's accuracy using the learning curve\cite{ref19}. This is based on an intuitive understanding that we can judge whether the model is reliable based on various indicator curves after the neural network has been trained for a period of time. The hierarchical representation believes that the search space would be enormous if the full neural network were to be explored. Zoph et al.\cite{ref20} devised NASNet, that by repeatedly building normal cells and reduction cells, drastically shrinks the search space. On the CIFAR-10 dataset, this cell-based network topology achieves good accuracy. This design also aids in the transformation of knowledge concurrently. The learned cell structure can help in more extensive image classification dataset and get state-of-the-art(SOTA) results on the object detection dataset. Although the inner structure of cell is learned, the outer linking of cells is predefined. This dramatically limits the diversity of cell structures, which lead to performance limitations. Therefore, for an efficient NAS algorithm, it is necessary to design a neural network structure that contains different cell structures in different parts. 

Except for performance prediction and hierarchical representation, weight sharing is used more frequently. The trained network is reused as often as feasible through weight sharing. The main stream of weight sharing is One-Shot Architecture Search\cite{ref15} that sample sub-architecture from SuperNet. only this SuperNet needs to be trained, since any sub-architecture is sampled from the SuperNet. The weights of all sub-architectures can be inherited from this SuperNet, which avoids training the sub-network from scratch and cutting the search time across several GPU days to few GPU hours. Although this weight-sharing method can effectively improve search efficiency, it still has limitations. Firstly, designing the SuperNet requires rich experiential knowledge. Secondly, the diversity of the network is limits since the network must be a subnetwork of the SuperNet. Benyahia et al.\cite{ref21} discovered that the shared weights of the previously trained sub-models will be overwritten during the training of subsequent sub-models when several sub-models are trained based on a single task through weight sharing. This is unfair when assessing network performance for the earlier sub-models. Additionally, the evaluation of the sub-model may deviate because different nodes in the SuperNet had varying degrees of weight training during the training phase. Specifically, when the subnet samples the poor nodes in SuperNet, its performance will not match its actual performance. Bender et al.\cite{ref22} randomly reset some operations to zero when training the supernet. By disabling path exit at the start of training and gradually raising the exit rate over time with a linear model, good results were obtained. However, their research indicates that the dropout rate has a significant impact on the model's predictive performance, which complicates the model training.

In this work, we present a two-stage search approach that includes a flexible encoding method and a weight inheritance strategy in order to employ evolutionary strategy to address the limits of the preceding. For weight inheritance during the search process, we also maintain a search space based on several cells. The following is a summary of the work's contributions.

\begin{enumerate}
\item{A two-stage search strategy is designed to discover more diverse network structures. In the first stage, our goal is to determine the framework of network structures by focuing on the network cell combination. In other words,  we roughly confirm the trend of the entire network structure at this stage. In the second stage, after the network structure is roughly fixed, we mainly conduct a global search for specific operations in the network. we perform cross-mutation and pruning operations on the nodes and edges inside the cell. In the end, an optimal network global structure is obtained.}
\item{A two-stage encoding method is proposed to express the entire network more concisely. first of all, different cells in the search space are encoded with the layer definition and layers connection. Then, the entire neural network structure is encoded.}
\item{A convenient and reliable weight inheritance method is proposed to accelerate the search process. We maintain an unchanged search space based on different cell combinations in the first search stage, therefore, the inherited cell can be found from original cell. During the search process, the nodes and connection methods within the cell remain unchanged and the weights of each new individual are inherited from the cell search space.}
\end{enumerate}

The remaining sections of this article are listed below. Section II contains background. The cell-based search space and two-stage encoding technique are described in Section III. In Section IV, the algorithm's framework will be explained. The algorithm's experimental design and results are provided in Sections V and VI. In Section VII, the conclusions are introduced.

\section{BACKGROUND AND RELATEDWORK}
ENAS is a crucial subfield in NAS that use evolutionary algorithms to find the ideal structure. To give the readers a thorough grasp of ENAS, we first discuss the most recent developments in this field. The fundamental elements of ENAS are then examined, with a focus on the application of encoding spaces and acceleration techniques.

\subsection{Evolutionary Neural Architecture Searchs}
Evolutionary algorithm is population-based gradient-free optimization algorithm. Inspired by biological evolution in nature, evolutionary algorithm typically includes fundamental operations such as gene encoding, population initialization, crossover and mutation operators and elite retention mechanism. Evolutionary computation, which exhibits self-organization, self-adaptation, and self-learning, is a well-developed global optimization technique with high robustness and adaptability. Evolutionary neural networks, also known as Neuroevolution\cite{ref23}, were developed decades ago by applying evolutionary algorithms to determine the structure, weights, and hyperparameters of neural networks. Similar to Neuroevolution, evolutionary algorithm-based neural network architecture search focuses on discovering the best neural network structure while still adhering to the broad principles of evolutionary algorithms. The algorithm process of ENAS first initializes a population of candidate neural network models via a predefined encoding strategy. A neural network is created for each member of the population, and it is trained using a specific dataset. Following training of the entire population, each person's performance is assessed on a validation set, and the test results are utilized to calculate their individual fitness values. Then, a selection strategy is employed to choose suitable parent individuals and offspring populations are generated through crossover and mutation operators. In order to continue optimizing iterations until terminative requirements are met, a new generation of populations is finally chosen using an elite retention method.

One of the early EvoNASs, LargeEvo\cite{ref24}, searches for improved Convolutional Neural Network (CNN) topologies using a genetic algorithm (GA)\cite{ref25}. A variable-length encoding system that employs a genetic algorithm to determine the ideal depth of CNN architectures was proposed by Sun et al\cite{ref26}. Besides genetic algorithms, other types of EC methods are also important components of ENAS algorithms, such as Evolution Strategies (ES), Particle Swarm Optimization (PSO)\cite{ref27}, Ant Colony Optimization (ACO)\cite{ref28} and Differential Evolution (DE)\cite{ref29}. ENAS is frequently employed in Convolutional Neural Networks (CNN) for image processing and Recurrent Neural Networks (RNN) for natural language processing because to the rapid development of deep learning. Numerous ENAS methods are utilized in the search for the ideal CNN architecture because CNN is widely applied in areas like object identification and image classification. The depth of the structure, connections between layers, and hyperparameters for all layers are the three fundamental components of CNN architecture optimization. In actuality, most ENAS methods take into account the above three aspects. For instance, the C. L. Philip Chen et al. proposed IDEA, an evolutionary algorithm for I Ching prediction based on macro search space, is primarily used to optimize the structural parameters of neural network architecture. The primary characteristic of RNNs is their recurrent connections, which allow the hidden layer nodes to receive output from earlier iterations of themselves in addition to input from the top layer. Therefore, certain ENAS approaches concentrate on the amount of times RNN should be unfolded\cite{ref31}, in contrast to CNN, which focuses on the number of layers and connections of the network. Additionally to the common CNN and RNN structures, there are also Deep Belief Networks (DBN)\cite{ref32} and Stacked Autoencoders (SAE). The main focus of ENAS approaches for DBN and SAE is the quantity of its constituent parts, specifically the Restricted Boltzmann Machines (RBM) that make up DBN and the Autoencoders (AE) that make up SAE\cite{ref33}.

In addition to searching for high-performance neural network models, ENAS algorithms need to consider multiple conflicting optimization objectives in practical application scenarios. These goals are particular to deployment settings. For instance, model size, latency, power consumption, and memory utilization are typically constrained in mobile or embedded systems. Specifically, it is necessary to optimize the floating-point operations (FLOPs) of the model \cite{ref34}, the latency of the model\cite{ref35}, the memory consumption of the model\cite{ref36} and the inference time, etc. Due to the frequent training and evaluation of several candidate models, these approaches nevertheless suffer from the drawback of high computational complexity.

\subsection{Encoding Space}
For ENAS algorithms, the network structures which are able to be searched depend on the search space, which is included within the encoding space. Therefore, a good encoding space can help find better network structures. It can be loosely classified into three types based on the basic unit used in the encoding space: layer-based coding space\cite{ref37},\cite{ref38},\cite{ref39},\cite{ref40}, block-based coding space\cite{ref41},\cite{ref42},\cite{ref43} and cell-based coding spaces\cite{ref44},\cite{ref45}. In addition, there are some ENAS methods that only focus on the connection between units, without considering the composition of basic units. This encoding space is called topology-based encoding space\cite{ref46},\cite{ref47}.

The basic unit in layer-based encoding space is the most fundamental layer of a neural network, such as the convolutional and pooling layers in CNN. Considering that a DNN contains numerous layers. For example, CNN usually consists of hundreds of basic layers. Therefore, the layer-based search space is very large and a lot of information needs to be encoded. This will also lead to the need for more time to search for a well-performing DNN. In addition, if only basic layers are used, it may not be possible to find excellent DNN structures, such as the skip-connection structure in ResNet\cite{ref6}.

Block-based encoding space is suggested as a solution to the aforementioned issues. The blocks contain blocks with certain topological relationships, such as remaining connections in ResBlock\cite{ref6} and strong connections in DenseBlock\cite{ref7}, and they are made up of different kinds of layers that are integrated into them. The basic units in the encoding space are these blocks with specified topologies, and deep neural networks are built by stacking these blocks. It is simpler to find an effective architecture in the block-based encoding space than in the layer-based encoding space because prior studies have demonstrated that these blocks have good performance. Additionally, because fewer parameters must be encoded when creating neural networks using block structures, encoding complexity and search time may be reduced. Some ENAS algorithms directly use these blocks\cite{ref8},\cite{ref48}, whilst other algorithms suggest various blocks for various uses. For instance, Chen et al.\cite{ref49} offered eight blocks encoded by 3-bit strings, such as ResBlock and InceptionBlock\cite{ref50}, and used Hamming distance to identify related blocks.

The block-based encoding space and the cell-based encoding space are related and the latter can be thought of as a specific instance of the former. The cell-based encoding space differs from the block-based encoding space in that the layers in the cell can be mixed more freely and there are typically just one or two cells present. By stacking repeated cells, ENAS algorithms create the architecture of neural networks.The micro part and the macro part of the cell-based DNN structure were split into two separate sections by Chu et al\cite{ref51}. In more detail, the macro part establishes the connections between cells while the micro part holds the characteristics in the cell. The macro part is defined by human skill, while the cell-based encoding space mostly concentrates on the micro part\cite{ref13},\cite{ref52},\cite{ref53}. This encoding space is widely used. For example, NAS-Bench-101\cite{ref52} looks for various arrangements of the cell's layers and connections before stacking the cells one at a time. Additionally, NASNet\cite{ref18} and DARTS\cite{ref54} look for two distinct cells called Normal and Reduction cells, and every stacked cell is linked to the two cells before it. The size of the encoding space is drastically reduced by the cell-based ENAS technique, which shifts the search burden from the entire network structure to one or two cell structures. However, since the macro part is defined by humans, this limits the algorithm's ability to search for more diverse structures. Moreover, Frachon et al.\cite{ref48} argued that the claim that the cell-based search space helps to achieve good DNN network structures has no theoretical basis.

\subsection{Acceleration method}
In the ENAS algorithm, the evolutionary operator is designed to preserve the structure of individuals to a degree, resulting in newly generated individuals that are structurally identical to their parents in certain parts. As such, weight inheritance from the parents is possible for these corresponding parts, thereby minimizing the need for complete training from scratch for many offspring individuals. This approach leads to significant savings in computing resources and time costs. It is worth noting that this technique was first adopted by earlier researchers approximately 20 years ago. In addition to weight inheritance, there is another unique ENAS method of reducing the number of populations to speed up evolution. However, this method may also lead to failure to explore the optimal network structure. To avoid this problem, Fan et al.\cite{ref44} adopted a strategy of dynamically reducing the population, dividing evolution into three stages and gradually reducing the population in each stage. A large population was used in the first stage to ensure global search capability. Population memory is another unique ENAS method. It works by recording previously seen architecture information in the population and reusing this information in the offspring. However, this method often consumes more memory than other ENAS methods.

\section{SEARCH SPACE AND CODING STRATEGY}

\begin{figure}
\centering
\includegraphics[width=80mm]{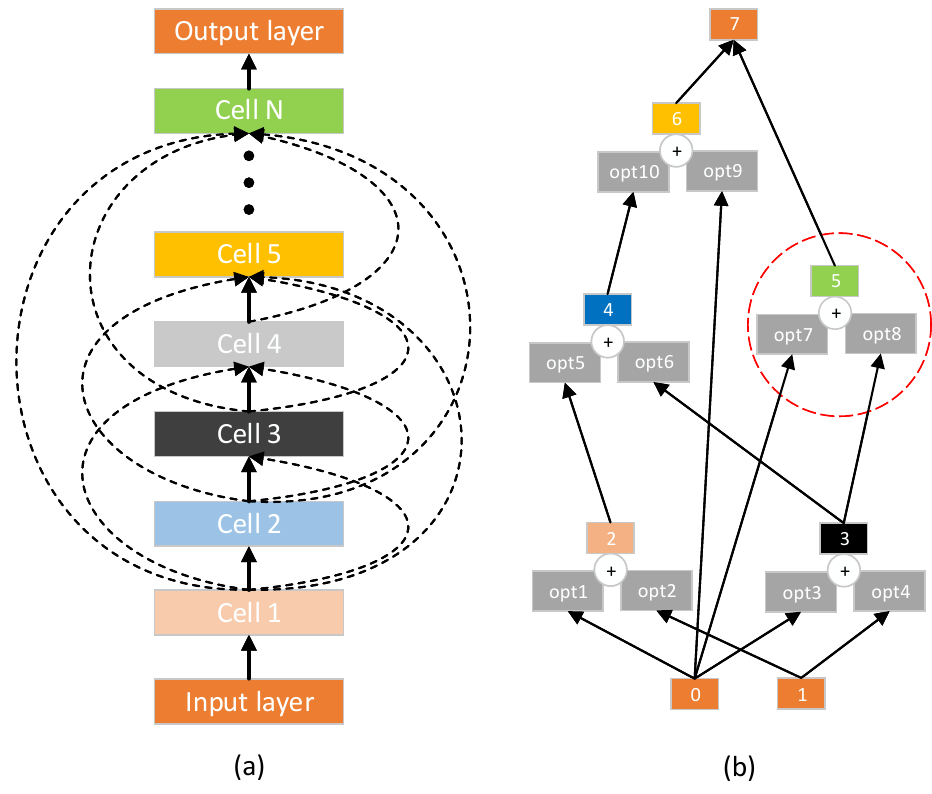}
\caption{Schematic diagram of the network structure and the internal structure of the cell.}
\label{fig1:env}
\end{figure}

\subsection{cell Search Space}
As shown in Fig.1(a), in our method, the network structure is a cell-based chain structure, mainly composed of an input layer, N cell layers and an output layer. For each cell in the network, we will number it from \{$cell_1$,$cell_2$,...,$cell_N$\}. The size of N determines the depth of the network. The value of N is not fixed, so the depth of the network is also not fixed. The determination of the value of N will be given in Section IV-B. In the network structure, in addition to the input and output layers, each cell will have two inputs, one of which comes from the previous cell, the arrow indicated by the solid line in the figure and the other input comes from another serial numbers except the previous cell In the previous cell, the arrow indicated by the dotted line in the figure. For example, for $cell_5$, its first input comes from $cell_4$ and the other input comes from one of the set \{$cell_1$, $cell_2$, $cell_3$\}. And for $cell_1$, its input is a copy of the input layer and the input layer. Compared with the way in which the input of the cell in AmoebaNet\cite{ref13} is the output of the first two cells, our connection method is more flexible and can search for more diverse network structures.

The cell can be seen as a tiny neural network made up primarily of nodes and directed edges, or as a directed acyclic graph. The smallest component of a cell is a node, and each node is a particular tensor (such as a feature map produced from a convolution operation). Input nodes, intermediate nodes, and output nodes are the three primary types of nodes in a cell. There are two input nodes among them that match to the two inputs of the network structure cell. Build the internal structure of the cell by include intermediary nodes and links between the input nodes and output nodes. Two inputs and at least one output are present in each intermediate node. The output must point to the output node if the node has only one output. To create an output node, all intermediary nodes with a single output are fused together. To make sure that the number of channels in each feature map stays constant while successfully fusing all of the input data, we often employ the add operation. Each directed edge represents a basic operation, such as convolution, pooling, etc. and represents an operation sampled from the operation space, through which the tensor of a node is transformed into the tensor of another node. It is worth noting that edges pointing to output nodes do not contain operations. Like NASNet\cite{ref18}, we set up thirteen basic operations and the specific operation space is shown in Table I. As shown in Fig.1(b), cell consists of 8 nodes (two output nodes $x_0$, $x_1$, five intermediate nodes $x_2$, $x_3$, $x_4$, $x_5$, $x_6$, one output node $x_7$) and 10 edge operations O={$o_1$,$o_2$,...,$0_10$} composition. Taking node 5 as an example, it is obtained by fusing the nodes $x_0$ and $x_3$ through the operation $o_7$ and $o_8$ after transforming them through the add operation. Since it has no output pointing to other intermediate nodes, it only has one output pointing to the output node. The output of node $x_5$ and the output of node $x_6$ are finally obtained by adding the output node $x_7$.

The size of the network is likewise unpredictable in our network architecture because of N. We need to encode the complete network since, in contrast to NASNet\cite{ref18}, we aim to find more diverse cells. As a result, the larger N is, the broader the network's search space and encoding space are, and the more complex the network's encoding is. We create a cell-based search space based on the aforementioned factors. The Normal Cell Search Space and the Reduction Cell Search Space are the two sections of the search space, respectively, as indicated in Table II. The Normal cell and the Reduction cell share the same structural characteristics, but the Normal cell only requires one operation step, whereas the Reduction cell requires two operations steps, doubling the number of channels. The cells in these two segments come from currently known cell modules with greater performance, such as modules searched by NASNet\cite{ref18} and DARTS\cite{ref54}, in order to reduce the difficulty of our search.

\begin{table}[]
    \caption{OPERATION SPACE}
    \vspace{5pt}
    \centering
    \begin{tabular}{m{3.50cm}<{\centering}m{1.75cm}<{\centering}m{1.25cm}<{\centering}m{0.5cm}<{\centering}}
        \hline
        Operation name & Operation type & Abbreviation & Code \\
        \hline
        identity & None & identity & 1 \\
        1×3 then 3×1 convolution & convolution & 1×3-3×1 & 2 \\
        1×7 then 7×1 convolution & convolution & 1×7-7×1 & 3 \\
        3×3 dilated convolution & convolution & dil3×3 & 4 \\
        3×3 average pooling & pooling & avg3×3 & 5 \\
        3×3 max pooling & pooling & max3×3 & 6 \\
        5×5 max pooling & pooling & max5×5 & 7 \\
        7×7 max pooling & pooling & max7×7 & 8 \\
        1×1 convolution & convolution & conv1 & 9 \\
        3×3 convolution & convolution & conv3 & 10 \\
        3×3 depthwise-separable conv & convolution & sep3×3 & 11 \\
        5×5 depthwise-separable conv & convolution & sep5×5 & 12 \\
        7×7 depthwise-separable conv & convolution & sep7×7 & 13 \\
        \hline       
    \end{tabular}
    \label{os1}
\end{table}

\begin{table}[]
    \caption{cell SEARCH SPACE}
    \vspace{5pt}
    \centering
    \begin{tabular}{m{4.0cm}<{\centering}m{1.35cm}<{\centering}m{1.25cm}<{\centering}m{0.4cm}<{\centering}}
        \hline
        cell name & cell type & Abbreviation & Code \\
        \hline
        AmoebaNet-A-Normalcell\cite{ref13} & Normalcell & AmoebaNC & 1 \\
        NASNet-A-Normalcell\cite{ref18} & Normalcell & NASNetNC & 2 \\
        DARTS-Normalcell\cite{ref54} & Normalcell & DartsNC & 3 \\
        CARS-H-Normalcell\cite{ref55} & Normalcell & CarsNC & 4 \\
        AmoebaNet-A-Reductioncell\cite{ref13} & Reductioncell & AmoebaRC & 5 \\
        NASNet-A-Reducitoncell\cite{ref18} & Reductioncell & NASNetRC & 6 \\
        DARTS-Reductioncell\cite{ref54} & Reductioncell & DartsRC & 7 \\
        CARS-H-Reductioncell\cite{ref55} & Reductioncell & CarsRC & 8 \\
        \hline       
    \end{tabular}
    \label{css1}
\end{table}

\begin{figure*}
\centering
\setlength{\abovecaptionskip}{0.cm}
\includegraphics[width=150mm]{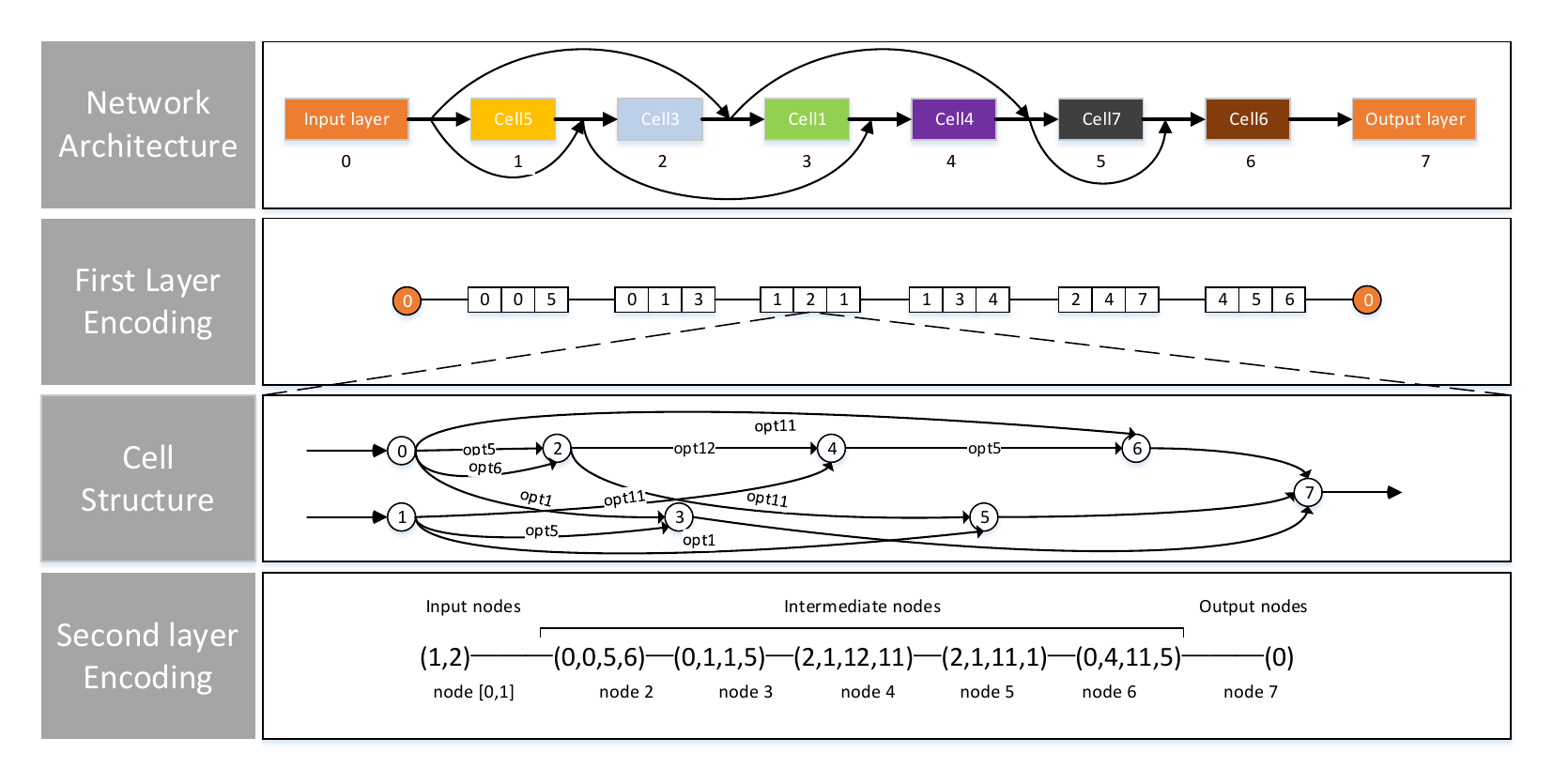}
\caption{Top: A network architecture composed of six cells. Second layer: the encoding of the network architecture. Third layer: a schematic diagram of the structure of $cell_1$ in the network architecture. Bottom: Code of $cell_1$.}
\label{fig1:env}
\end{figure*}

\subsection{Two-Stage encoding Method}
The two-stage encoding method designed in this paper will be introduced in detail below. Our  two-stage encoding method and two-stage search strategy are complementary. In the rough search stage, since the cell is fixed, the cell can be encoded directly without considering the internal structure of the cell. Firstly, the eight cells in the cell search space are encoded with real numbers and the encoding sequence numbers are from 1-8. Among them, the numbers 1-4 are Normal cells and the numbers 5-8 are Reduction cells. Since each cell has two inputs, two number bits are required to represent the two inputs of the cell. The number in the serial number indicates that the input of the cell is the output of a specific layer in the network. For example, if the two number bits of a cell are 0 and 1, respectively, it means that the first input of the cell is the output of the input layer in the entire network and the second input is the output of the first cell in the network. After the rough search is completed, enter the fine search stage. The fine search stage is mainly to adjust the internal structure of the cell, so the internal structure of the cell needs to be encoded. For the input node of cell, you can directly use the number bit representation of the first layer, such as (0,1). For intermediate nodes, we use a 4-tuple ($x_i$, $x_j$, $o_i$, $o_j$) for encoding. In this scheme, the first two integers represent input node 1 and input node 2. The last two integers represent the operation to be performed by input node 1 and the operation to be performed by input node 2, respectively. For example, for the quadruple (0,1,1,2), the input nodes are 0 and 1. For the input node 0, the operation (identity) with the sequence number 1 in the operation space needs to be performed. For the input node 2, It is necessary to perform the operation with the sequence number 2 in the operation space (1x7 then 7x1 convolution). Finally, the output node is directly represented by the real number 0.

This two-stage encoding method is suitable for our two-stage search strategy. In the rough search stage, using the first layer of coding to represent the network can not only represent the cell-based network well but also simplify the coding difficulty and provide a more convenient operation method for the cross-mutation operation. This helps us explore more diverse networks. In the fine search stage, the second layer of coding is more concerned with changes in the internal structure of the cell, which helps to find the optimal structure.

\begin{algorithm}
\caption{Framework of TS-ENAS}\label{alg:alg1}
\KwIn {population size $S$.}
\KwOut {Best individuals.}
$P_{0}\leftarrow$Initialize the population in the cell search space with two-layer encoding strategy\;
$t\leftarrow$0\;
\While{termination criterion is not satisfied}{
Evaluate the fitness of individuals in $P_{t}$ by using the proposed cell-based weight inheritance\;
$Q_{t}\leftarrow$Generate offspring from $P$ using genetic operators\;
$P_{t+1}\leftarrow$Environmental selection from $P_{t}\cup Q_{t}$ by using the proposed strategy\;
$t\leftarrow$$t+1$\;
}
$P_{t}\leftarrow$Select the best $S$ individuals from $p_{t}$\;
\While{termination criterion is not satisfied}{
$Q_{t}\leftarrow$Generate offspring with the designed genetic operators from $P_{t}$\;
Using the proposed cell-based weight inheritance to evaluate the fitness of newly generated individuals\;
$P_{t+1}\leftarrow$Select the best S individuals from $P_{t}\cup Q_{t}$\;
}
Select the best individual from $P_{t}$ and decode it into the corresponding deep neural network.
\label{alg1}
\end{algorithm}

\section{PROPOSED ALGORITHM}
In this section, the proposed Two-Stage Evolutionary Search for Neural Network Architectures With cell Search Space(TS-ENAS) is introduced in detail.

\subsection{Algorithm Overview}
The proposed TS-ENAS method's framework is described in Algorithm 1. A two-stage evolutionary search technique is used in the TS-ENAS approach. First, the population is initiated using the two-layer gene coding technique that has been suggested (line 1). The first phase of evolution then begins (lines 3–8). This stage's major goal is to identify multiple network architectures made up of cells that excel at the intended activities. We call this stage the rough search stage. The second stage of evolution then takes over. at the network structure that was first searched, the internal structure of the cell is adjusted at this stage. To keep the algorithm from entering local optimization is the major goal. The fine search is the name of the second step. The best person is then chosen and encoded into the appropriate deep neural network (line 15) for post-training.

In the rough search stage of the two-stage evolutionary process, all individuals are first evaluated according to the proposed cell-based weight inheritance method. New offspring are generated through specific cross-mutation operators and then representatives are selected from existing individuals and newborn individuals to form the next generation of the population, participate in subsequent evolution until the termination conditions of this stage are met and finally, the optimal $S$ individuals from the population are selected to participate in the second stage of evolution. In the evolution process of the fine search stage, all individuals in the population are first evolved by the designed, evolutionary operator and then the fitness of the newly produced individuals is evaluated using the same weight inheritance method. Next, representatives are selected from existing and newborn individuals to form the next generation of the population to participate in subsequent evolution. Until the termination condition is met. In the following sections, we will describe the key steps in Algorithm 1 in detail.

\begin{algorithm}
\caption{Rough Search Phase}\label{alg:alg1}
\KwIn{population size $P$,the maximum number of iterations $T_0$,crossover rate $\gamma$,mutation rate $\beta$,$S$}
\KwOut{Best $S$ individuals.}
$P_t\leftarrow$A population of size $P$ is initialized by \textbf{the proposed first-layer encoding strategy}\;
$W_{super}\leftarrow$Build a cell-based SuperNet\;
Training SuperNet $W_{super}$\;
$t\leftarrow$0\;
\While{$t<T_0$}{
Use \textbf{cell-based weight inheritance} to initialize weights and evaluate the fitness value of individuals in $P_t$\;
$Q_t\leftarrow$$\emptyset$\;
\While{$\lvert Q_{t} \rvert < P$}{
$p_1,p_2\leftarrow$Select two parent individuals from $P_t$\;
$q_1,q_2\leftarrow$Use\textbf{the designed crossover operation}to generate two offspring individuals with probability $\gamma$\;
$p_3\leftarrow$Select an individual from $p_t$\;
$c\leftarrow$Random generated a number between[0,1]\;
\eIf{$c=0$}{
$q_3\leftarrow$Use \textbf{cell mutation} to mutate $p_3$ with a probability of $\beta$\;
}{
$q_3\leftarrow$Use \textbf{connect mutation} to mutate $p_3$ with a probability of $\beta$\;
}
$Q_t\leftarrow$$Q_t\cup q_1\cup q_2 \cup q_3$\;
}
Evaluate the fitness of individuals in $Q_t$\;
$P_{t+1}\leftarrow$Select $P$ individuals from $Q_t\cup P_t$ using \textbf{the environment selection strategy}\;
$t\leftarrow$$t+1$\;
}
\textbf{Return} The best $S$ individual.
\label{alg1}
\end{algorithm}

\subsection{Rough Search Phase}
In the rough search stage, the algorithm mainly focuses on the entire network and its main purpose is to find the optimal combination, quantity and connection mode of each cell in the network. The primary contributions of this research are highlighted in italic and bold in Algorithm 2, which also includes a list of the pseudocode for the main coarse search components. A population of size $P$ is first initiated using the suggested coding scheme. Then, construct a SuperNet according to the cell search space and use the training set to train the SuperNet. Then, enter the evolution stage. In the evolution stage, the weights of the individual $P_t$ in the population are initialized using the cell-based weight inheritance method and then these individuals are trained on a small scale. After the training is completed, the fitness of all individuals is evaluated on the validation set $D_{valid}$. Then, new offspring individuals are generated using the proposed crossover mutation operation. Repeat the crossover mutation operation until $P$ offspring individuals are generated, then train on a small scale and evaluate the fitness of all individuals in $Q_t$ and generate the next generation of parental population $P_{t+1}$ through the proposed environment selection strategy. Finally, select $S$ individuals to enter the next stage of evolution. Its main components will be described in detail below.

\begin{algorithm}
\caption{Population Initialization}\label{alg:alg1}
\KwIn{$N_{max}$, $N_{min}$,population size $P$}
\KwOut{Initialize population $P_0$}
$P_0\leftarrow$$\emptyset$\;
\For{$i\leftarrow 0$ \KwTo $P$}{\label{forins}
$N_{i}\leftarrow$Randomly generate an integer from [$N_{max}$,$N_{min}$]\;
$p_{i}\leftarrow$$\emptyset$\;
$p_{i}\leftarrow$$p_i\cup 0$\;
$k\leftarrow$0\;
\For{$j\leftarrow 1$ \KwTo $N_i$}{\label{forins}
$a_{1}\leftarrow$$j-1$\;
$a_{2}\leftarrow$Randomly generate two integers from [0,j-1]\;
\eIf{$k<\frac{n_i}{2}$}{
$a_{3}\leftarrow$Randomly generate two integers from [1,8]\;
\If{$a_{3}\in[5,8]$}{
$k\leftarrow$$k+1$\;
}
}{
$a_{3}\leftarrow$Randomly generate two integers from [1,4]\;
}
$p_{i}\leftarrow$$p_i\cup a_1\cup a_2\cup a_3$\;
}
$P_{0}\leftarrow$$P_o\cup p_i$\;
}
\textbf{Return} $P_0$\;
\label{alg1}
\end{algorithm}

\subsubsection{Population Initialization}
Before the population is initialized, it is first necessary to limit the size of the network so as not to search for a network that is too large or too small. We limit the maximum and minimum depth of the network by two values: $N_{max}$ and $N_{min}$. The size of each individual in the population cannot exceed this range. Algorithm 3 shows the main steps of population initialization. When the population is initialized, we first randomly generate a value $n_i$ in [$N_{max}$,$N_{min}$], which represents the number of cells in this individual. Then, randomly select an integer $a_1$ in [0, $j-1$) as the first serial number of the current cell. It is worth noting that if the number $j$=1 of the current cell, the first number bit is $j-1$. Then, select the previous connection $j-1$ of the current cell as the second serial number $a_2$ of the current cell. Then, randomly generate a cell code $a_3$. Generally speaking, we agree that the number of Reduction cells in a network is smaller than the number of Normal cells, so we need a counter k to record the number of Reduction cells in the network. When $k<\frac{n_i}{2}$, the selection range of $a_3$ is the entire cell Search space, when $k>=\frac{n_i}{2}$, the selection range of $a_3$ is the Normal cell search space. Finally, repeat the same method to generate the initial population.

\subsubsection{Use cell-based weight inheritance for weight initialization and fitness evaluation}
As we all know, network training is very time-consuming, so it is necessary to choose an appropriate way to avoid heavy training tasks. In our method, since the first stage of the search is based on the cell, the cell can be regarded as a complete small network because the internal structure of the cell will not be destroyed during the evolution process. Therefore, each cell, its weight can be inherited. That is, the weight of the cell circulates between the parent and the child. Specifically, first stack all the cells to form a SuperNet $W_{super}$, train the SuperNet on the training set and then record the weights of each cell in the SuperNet respectively. When a new individual appears in the population, the recorded cell weight can be directly reused as the initialization weight of the entire network. Of course, the inherited weights cannot be directly used for the performance test of the network because it does not represent the performance of the network, so it needs to be trained on a small scale (generally 10-20 Epochs) before they can be used for the final test. Finally, the test results are used as the individual fitness value for environment selection. After the population has been updated for a certain number of generations, the weights of the recorded cells also need to be updated. The specific update steps are as follows:

\begin{enumerate}[Step 1:]
    \item Select the best individual in this generation population.
    \item Use the weight of the cell in this individual to update the weight of the cell that has not been updated in the search space.
    \item Determine whether all the cells in the search space have been updated.
    \item If so, the loop ends. Otherwise, select the suboptimal individual in the population and perform steps 2-4 until all the cells in the search space are updated.
\end{enumerate}

\subsubsection{Offspring Generation}

\begin{figure*}
\centering
\setlength{\abovecaptionskip}{0.cm}
\includegraphics[width=150mm]{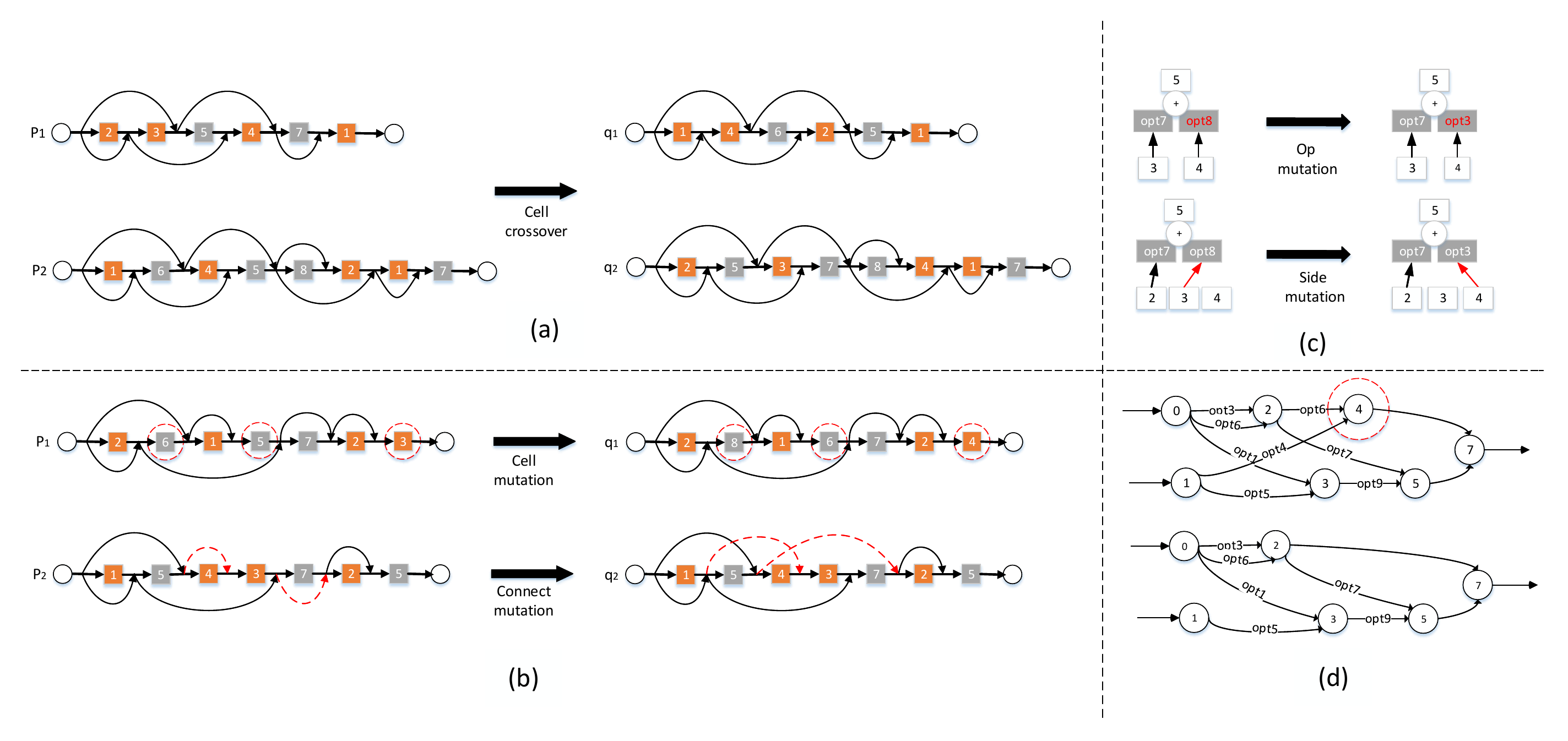}
\caption{Illustration of the proposed crossover, mutation and pruning operators. (a) is the intersection method of the rough search stage. (b) is the mutation mode in the rough search stage. (c) is the mutation method in the fine search stage. (d) is the pruning operation in the fine search stage.}
\label{fig1:env}
\end{figure*}

In an evolutionary algorithm, given an initial population, a new offspring population is generated through a crossover mutation operator. In general, the crossover operation performs a local search, while the mutation operation performs a global search. The two work together to produce better-performing offspring. As shown in Fig.3(a), in our method, the crossover adopts the way of multi-point crossover. It is worth noting that since the individual $N$ values in the population are different, for the two selected parent individuals $p_1$ and $p_2$, we will select the individual with the smaller N value as the reference individual and then sequentially compare the cells are interleaved until all cells in the benchmark are traversed. In this process, we will follow the principle that the cell category remains unchanged. Specifically, the Normal cell in $p_1$ will only be exchanged with the Normal cell in $p_2$ and the Reduction cell in $p_1$ will only be exchanged with the Reduction cell in $p_2$. When the number of Reduction cells in $p_2$ is not enough for a complete exchange, we will skip the current cell.

As shown in Fig.3(b), the mutation operations proposed in this paper are mainly divided into two types. The first is the mutation of the cell. Specifically, multiple mutation points are randomly selected in the individual at first and the mutation points at this time are codes representing cells in the chromosome. For a selected mutation point, another cell can be randomly selected in the cell search space for replacement. Similarly, this replacement also needs to follow the principle that the cell category does not change. That is, Normal cells can only be replaced with Normal cells and Reduction cells can only be replaced with Reduction cells. The second is the variation of the connection method. The mutation point at this time is the coding bit representing the first input in the chromosome. Similarly, the mutation method is to randomly generate another connection that meets the requirements to replace the old connection.

\begin{algorithm}
\caption{Fine Search Stage}\label{alg:alg1}
\KwIn{the population $P_s$ obtained by rough search,the maximum number of iterations $T_1$}
\KwOut{best individual $p_{best}$}
The second layer of encoding for the population $P_s$\;
$t\leftarrow$0\;
\While{$t<T_1$}{
\For{each individual $p$ in $P_s$}{\label{forins}
$p_1\leftarrow$Use \textbf{Node mutation or Edge mutation} for individual $p$ to generate new individuals\;
$p_2\leftarrow$Use \textbf{Node pruning} for individual $s$ to generate new individuals\;
Evaluate the fitness values of individuals $s$ ,$p_1$ and $p_2$\;
$p\leftarrow$Select the best individual in set \{$p$,$p_1$,$p_2$\}\;
}
$t\leftarrow$$t+1$\;
}
\textbf{Return} $p_{best}$\;
\label{alg1}
\end{algorithm}

\subsubsection{Environmental Selection}
In the evolutionary algorithm, for each generation of individuals, not only their fitness but also their distribution should be considered. Specific to our rough search stage, when selecting the next generation of individuals, we not only need to consider the accuracy rate but also the distribution of the size of $N$ to prepare for the next stage of the search. First, we set a threshold $\alpha$ for the difference between the accuracy rates of two individuals. If the absolute value of the difference between the accuracy rates of two individuals is less than $\alpha$, the two individuals are considered to be equally good. That is, they are at the same level. For all individuals in the population, the hierarchy is divided according to $\alpha$. Then put into the next generation population sequentially according to the size of the hierarchy. When putting into the next level will exceed the limit of the population size of the next generation, evaluate the density value of the individuals in the next level and select individuals with low-density values to enter the next generation population until the population reaches the upper limit. The specific evaluation method of the density value is: for the individual in the next level, count the number of individuals that have been placed in the next generation population with the same $N$ value and this value is the density value.

\subsection{Fine Search Stage}
In the fine search stage, the algorithm mainly focuses on the cell part of the network, with the purpose of adjusting the already searched network structure to find the global optimal network structure. The specific search steps are given in Algorithm 5. The input of the fine search stage is the $S$ individuals selected in the rough search stage. For each individual, we generate two new individuals based on mutation and pruning operations. Then, train and evaluate the two newly generated individuals. Compare the new individual with the old individual and select the best individual to enter the next generation of evolution. Repeat the above operation until $S$ individuals of the next generation are selected. Repeat iterations until constraints are met. Finally, output the best individual.

In this stage, the main operations to generate offspring are mutation and pruning. As shown in c in Fig3.(c), we refer to the mutation operation in AmoebaNet and divide the mutation into two types. The first is edge replacement. The specific operation is to randomly select another operation from the search space to replace a certain edge operation in the cell. The second is the replacement of the input object of the node. Select a node in the cell and replace the input node of the node with another node. In the actual application process, we will apply these two mutation methods to an individual at the same time. Specifically, for each cell in an individual, there will be an equal probability of performing the first or second mutation until all the cells complete the mutation operation. In addition to the above two mutation operations, there is also a pruning operation for generating new cell structures. As shown in Fig3.(d). Specifically, for the selected cell, we will randomly delete a certain node in the cell. The main purpose of the pruning operation is to discover more network structures of different depths and increase the diversity of the network. In coding, we will set both the input of the deleted node and the edge operation to zero.

\section{EXPERIMENTAL DESIGN}
To verify the performance of the proposed algorithm, we design and execute a series of experiments on selected image classification benchmark datasets and compare the results with several current popular methods. Below, we first briefly introduce these benchmark datasets and the baselines for comparison. Then, the implementation details of the proposed method are presented.

\subsection{Benchmark Datasets}
In these experiments, we use three commonly used benchmark classification tasks Fashion-MNIST\cite{ref56}, CIFAR10\cite{ref11} and CIFAR100\cite{ref11}. As a substitute for the MNIST handwritten dataset, Fashion-MNIST increases the difficulty of feature extraction to a certain extent. The Fashion-MNIST dataset contains frontal images of 70,000 different clothing items from 10 categories and each image is preprocessed into a grayscale image with a resolution of 28x28. The CIFAR-10 dataset is a collection of images commonly used to train deep neural networks. The CIFAR-10 data set contains 10 categories and each category contains 6000 images, of which 5000 are used as training data, 1000 are used as test data and each picture is fixed at 32*32 pixels. CIFAR-100 is based on the category of CIFAR-10 and the dataset contains 20 superclasses and each superclass is further subdivided into 5 categories. Each class contains 600 images, including 500 training images and 100 testing images. Compared with CIFAR-10, the number of pictures for each class is reduced by 10 times, which increases the difficulty of convergence during training. At the same time, in order to test the performance of the algorithm on large-scale datasets, we also conducted experiments on the ImageNet\cite{ref57} dataset. It should be noted that due to computational cost constraints, our network is searched on a small dataset and then migrated to the ImageNet dataset.

\subsection{Baselines}
To verify the effectiveness and superiority of the proposed algorithm, we compared it with three different types of peer competitors. The first category is a hand-designed neural network architecture, many of which are selected from the progress ranking provided by the homepage of the dataset, mainly including VGG\cite{ref5}, ResNet\cite{ref6}, DenseNet\cite{ref7}, 2C1P2F+Dropout \cite{ref26}, 2C1P\cite{ref26}, 3C2F\cite{ref26}, 3C1P2F+Dropout\cite{ref26} and XgBoost.The second type of CNN architecture is mainly designed by non-ENAS algorithms, mainly including NASNet\cite{ref18}, DARTS\cite{ref54}, ENAS\cite{ref15} and PNAS\cite{ref16}. The third category is the CNN architecture designed by the ENAS algorithm, mainly AmoebaNet-A\cite{ref13}, AE-CNN\cite{ref58}, NSGANet\cite{ref59}, CNN-GA\cite{ref60} and Large-scale Evo\cite{ref24}, etc.

\subsection{Parameter Settings}
In our algorithm, except for $N_{max}$ and $N_{min}$ about the cell depth, all other parameters are selected according to the agreement of the evolutionary algorithm and the deep learning community. Specifically, the population size and total generation number of the first stage are both set to 100 and the population size and total generation number of the second stage are set to 30 and 50, respectively. The probabilities of crossover and mutation of the algorithm are 0.8 and 0.2, respectively. In addition, according to the agreement of the deep learning community, we will randomly sample 10\% of the data from the training images as the validation data set. The SuperNet training Epoch is set to 100 and the small-scale training Epoch of each individual is set to 10. The gradient descent strategy uses the SGD algorithm and sets the weight decay factor to $3*10^{-4}$ and the SGD momentum to 0.9.

\begin{table*}[]
    \caption{Comparison of TS-ENAS and hand-designed networks in terms of test classification error (Top1 error), number of parameters (parameter), training epoch (epochs) and Parameter number ratio on Fashion-MNIST benchmark datasets}
    \vspace{5pt}
    \centering
    \begin{tabular}{m{3.25cm}<{\centering}m{2.25cm}<{\centering}m{2.25cm}<{\centering}m{1.0cm}<{\centering}m{3.00cm}<{\centering}}
        \hline
        Model & Top1 classification error(\%) & \#parameters(M) & Epochs &  Parameter number ratio (with TSE-Net(V1))\\
        \hline
        2C1P2F+Dropout\cite{ref26} & 8.4 & 3.27 & 300 & 3.80x\\
        3C2F\cite{ref26} & 9.3 & - & - & -\\
        3C1P2F+Dropout\cite{ref26} & 7.4 & 7.14 & 150 & 8.30x\\
        GRU+SVM+Dropout\cite{ref61} & 10.3 & - & \textbf{100} & -\\
        GoogleNet\cite{ref50} & 6.3 & 101 & 200 & 117.44x\\
        AlexNet\cite{ref2} & 10.1 & 60 & 300 & 69.77x\\
        SqueezeNet-200\cite{ref62} & 10.0 & \textbf{0.5} & 200 & \textbf{0.58x}\\
        VGG16\cite{ref5} & 6.5 & 26 & 200 & 30.23x\\
        ResNet18\cite{ref6} & 5.1 & 11.19 & 200 & 13.01x\\
        DenseNet-BC\cite{ref7} & 4.6 & 0.77 & \textbf{100} & 0.90x\\
        XgBoost & 10.2 & - & 500 & -\\
        \hline
        EvoCNN\cite{ref26} & 5.5 & 6.68 & \textbf{100} & 7.77x\\
        ours(V1) & \textbf{4.4} & 0.86 & \textbf{100} & 1.00x\\
        \hline       
    \end{tabular}
    \label{css1}
\end{table*}

\section{EXPERIMENTAL RESULTS AND ANALYSIS}
In this section, we present the experimental results of the algorithm on three benchmark classification datasets. Specifically, in comparison with the manually designed network, it mainly includes four aspects: TOP1 classification error, number of parameters, training rounds and the ratio of the number of parameters. In comparison with peers, we also adopted GDs (GPU days), which is the time spent on the GPU, to measure the efficiency of the algorithm. At the same time, in order to verify the effectiveness of the two-stage strategy, we observed the accuracy of each generation of the optimal individual in the TS-ENAS algorithm population and the evolutionary trajectory of the entire population. Then, we analyzed the impact of the weight inheritance strategy on the algorithm from two aspects of accuracy and time loss. Finally, we also analyze the effectiveness of the proposed crossover, mutation and pruning operations.

\subsection{Results}
Tables III shows the comparison results of the algorithm on the Fashon-MNIST dataset and many excellent networks. The results of the algorithm on the CIFAR10\cite{ref11} and CIFAR100\cite{ref11} datasets are given in Tables IV. In Table V, we present the results of the algorithm migration to ImageNet. Best accuracy and highlighted in bold. The model searched by our algorithm is called TSE-Net. Note that most of the experimental results of competitors in the table are extracted from their original papers.

\subsubsection{Results on Fashion-MNIST}
As can be seen from Tables III, the algorithm is compared with 11 advanced hand-designed networks in Top1 classification error, weight super-parameter scale and training rounds and the network architecture EovCNN\cite{ref26} based on evolutionary strategy design is also added. The experimental results show that, with fewer training rounds, the architecture searched by the proposed search strategy proposed in this paper has a parameter size of 0.86 M. The classification error of Top1 in the comparison algorithm is 4.43\% and among all the algorithms participating in the comparison, the classification error is 4.43\%. Is optimal. Under the same training rounds, the EovCNN model parameter size is 7.77 times that of the proposed architecture, but the classification error is 1.1\% higher, which further illustrates the effectiveness of the search strategy proposed in this paper. It should be noted that the training round adopts the smallest round among all compared algorithms, but the algorithm can further converge in more rounds during the experiment, so the current result is not the best result of the algorithm. According to the experimental results, it can be shown that on the Fashion-MNIST dataset, the proposed algorithm has a strong competitive advantage in terms of accuracy and model size.

\subsubsection{Results on CIFAR10 and CIFAR100}
As shown in Tables IV, compared with the manually designed network, the algorithm has obvious advantages in the classification error indicators of CIFAR10 and CIFAR100. Compared with DenseNet-BC\cite{ref7}, the error of our algorithm on CIFAR10 is reduced by nearly 1\%, while the error on CIFAR100 is reduced by nearly 5\%. Compared with the non-evolutionary NAS algorithm, the proposed algorithm is still ahead of most algorithms in terms of accuracy and the search time also has a greater advantage. For example, compared with the gradient-based algorithm DARTS\cite{ref54} with lower time cost, our algorithm is more dominant in classification error (CIFAR10: 2.43\%<3.00\%) and consumes less search time (0.9GDs<1.5GDs). At the same time, compared with similar ENAS algorithms, the accuracy and parameter scale of the algorithm also have obvious advantages. Although the proposed algorithm LEMONADE\cite{ref69} is close to the search architecture in accuracy, it has achieved a large lead in search time (0.9GDs<13.1GDs). For example, compared with AmoebaNet, the search efficiency is improved by about 3500 times.

\begin{table*}[]
    \caption{COMPARISON BETWEEN TS-ENAS AND HAND-DESIGNED NETWORK AND OTHER NAS AND ENAS IN TERMS OF THE TEST CLASSIFICA TION ERROR (Top1 error), THE NUMBER OF PARAMETERS (P) and THE SEARCH COST (IN GDS) ON THE CIFAR10 and CIFAR100 BENCHMARK DATASETS}
    \vspace{5pt}
    \centering
    \begin{tabular}{m{2.90cm}<{\centering}|m{1.25cm}<{\centering}m{1.25cm}<{\centering}m{1.0cm}<{\centering}|m{1.25cm}<{\centering}m{1.25cm}<{\centering}m{1.00cm}<{\centering}|m{2.00cm}<{\centering}}
        \hline
        \multirow{2}{*}{Model}    & \multicolumn{3}{c|}{CIFAR10} & \multicolumn{3}{c|}{CIFAR100} & \multirow{2}{*}{Search method} \\ 
        \cline{2-7} 
        &\multicolumn{1}{c}{Top1 error(\%)} &\multicolumn{1}{c}{\#P(M)} &\multicolumn{1}{c|}{GDs} & \multicolumn{1}{c}{Top1 error(\%)} & \multicolumn{1}{c}{\#P(M)}& \multicolumn{1}{c|} {GDs}\\
        \hline
        DenseNet-BC\cite{ref7} & 3.46 & 25.26 & - & 22.28 & 25.26  & - & manual\\
        ResNet (depth=110)\cite{ref6} &6.43&1.7&-&27.81&1.7& - & manual\\
        ResNet(depth=1202)\cite{ref6} &7.93&10.2&-&25.16&10.2& - &manual\\
        VGGNet\cite{ref5} &6.66 & 20.4 & - & 28.05 & 20.04 &- & manual\\
        All-CNN\cite{ref60} &7.25 & \textbf{1.3} & - &33.71 & \textbf{1.3}& - &manual\\
        \hline
        DARTS\cite{ref54} & 3.00 & 3.3 & 1 & 17.54 & 3.3 & 1 & gradient\\
        ENAS\cite{ref15} & 2.89 & 4.6 & 0.5 & 19.43 & 4.6 & 0.5 & gradient\\
        SNAS\cite{ref63} & 2.98 & 2.8 & 1.5 & 20.09 & 2.8 & 1.5 & gradient\\
        PC-DARTS\cite{ref64} & 2.57 & 3.6 & 0.1 & 17.11 & 3.6 & 0.1 & gradient\\
        NASNet(A)\cite{ref18} & 2.65 & 3.2 & 1800 & 17.81 & 3.2 & 1800 & RL\\
        Proxyless NAS\cite{ref65} & \textbf{2.08} & 5.7 & 1500 & - & -& - & RL\\
        BlockQNN\cite{ref66} & 3.30 & 6.1 & 90 & 17.05 & 6.1 & 90 & RL\\
        RENA\cite{ref67} & 3.87 & 3.4 & - & - & - & - & RL\\
        MdeNAS\cite{ref68} & 2.55 & 3.8 & \textbf{0.16} & 17.61 & 3.8 & \textbf{0.16} & MDL\\
        PNAS\cite{ref16} & 3.41 & 3.2 & 225 & 17.63 & 3.2 & 225 & SMBO \\
        \hline
        AmoebaNet-A\cite{ref13} & 3.12 & 3.1 & 3150 & 18.93 & 3.1 & 3150 & EC\\
        AE-CNN\cite{ref58} & 4.70 & 2.0 & 27 & 22.40 & 5.4 & 36 & EC\\
        NSGANet\cite{ref59} & 3.85 & 3.3 & 8 & 20.73 & 3.3 & 8 & EC\\
        Large-scale Evo\cite{ref24} & 5.40 & 5.4 & 2750 & 23.00 & 5.4 & 2750 & EC\\
        LEMONADE\cite{ref69} & 2.58 & 13.1 & 90 & - & - & - & EC\\
        CNN-GA\cite{ref60} & 3.34 & 2.9 & 35 & 20.53 & 4.1 & 40 & EC\\
        SI-EvoNAS\cite{ref70} & 2.69 & 1.84 & 0.458 & \textbf{15.70} & 3.32 & 0.813 & EC\\
        \hline
        ours & 2.43 & 2.98 & 0.9 & 17.31 & 3.35 & 0.9 & EC\\
        \hline       
    \end{tabular}
    \label{css1}
\end{table*}

\subsubsection{Results on ImageNet}
Due to the high computational cost of the NAS algorithm, especially the search efficiency on large-scale data sets is very low. For example, on the ImageNet dataset, the time required for proxyless NAS\cite{ref65} search is 8.3GDs. Therefore, as a compromise, most NAS algorithms suggest searching for optimal networks on small datasets and transferring the found networks to large-scale tasks. However, the network found by this approach is not the optimal network for the target task. In our work, we migrated the optimal network found on the CIFAR10 dataset to the ImageNet dataset and the specific results are shown in Table V. The top 1 accuracy of our searched architecture is 75.67\%. Compared with peers, this accuracy rate has no obvious advantage. But our architecture is more optimal in terms of the number of parameters (3.67M) and the time cost of searching (0.9GDs).

\begin{table}[]
    \caption{Comparison of TS-ENAS and other NAS in terms of classification error (error), number of parameters (P) and consumed search costs (GDs) on the IMAGENET benchmark dataset}
    \vspace{5pt}
    \centering
    \begin{tabular}{m{2.5cm}<{\centering}|m{1.0cm}<{\centering}|m{0.7cm}<{\centering}|m{0.7cm}<{\centering}|m{1.1cm}<{\centering}}
        \hline
        \hline
        Model & Top1 error(\%) & \#P(M) & GDs &Search dataset\\
        \hline
        AmoebaNet\cite{ref13} & 24.3 & 6.4 & 3150 & CIFAR10\\
        \hline
        NASNet-A\cite{ref18} & 26.0 & 5.36 &1800&CIFAR10 \\
        \hline
        PNAS\cite{ref16} & 25.8 & 5.1 & 225 & CIFAR10 \\
        \hline
        SNAS(mild)\cite{ref63} & 27.3 & 4.3 & 1.5 & CIFAR10\\
        \hline
        DARTS\cite{ref54} & 26.7 & 4.7 & 4.0 & CIFAR10 \\
        \hline
        RENAS\cite{ref71} & 24.3 & 5.36 & 1.5 & CIFAR10\\
        \hline
        PDARTS\cite{ref72} & 24.7 & 5.1 & 0.3 & CIFAR100\\
        \hline
        PC-DARTS\cite{ref64} & 24.2 & 5.3 & 3.8 & ImageNet \\
        \hline
        MnasNet-A3\cite{ref73} & 23.3 & 5.2 & - & ImageNet \\
        \hline
        Proxyless NAS\cite{ref65} & 24.9 & 7.1 & 8.3 & ImageNet \\
        \hline
        ours & 24.33 & 3.67 & 0.9 & CIFAR10 \\
        \hline
        \hline
    \end{tabular}
    \label{css1}
\end{table}

\subsection{Analysis of the effectiveness of the Two-Stage strategy}
Most of the existing NAS algorithms search for two kinds of cells on the target task: Normalcell and Reductioncell. Then these two kinds of cells are stacked in a specific order to build a complete network. Therefore, the approximate structure and depth of the network may be estimated as long as the number of nodes inside the cell and its stacking times are known. The entire search procedure is split into two parts by the TS-ENAS algorithm. It primarily looks for connections and combinations between cells in the first step. The second stage involves optimizing the cell's interior in order to locate the best network. The final experimental findings demonstrate that the method performs well on both the small Fashion-MNIST dataset and the broader ImageNet dataset. On the CIFAR10 and Fashio-MNIST datasets, we simultaneously assess the accuracy of the ideal person in each generation of the search process to confirm the efficacy of the two-stage technique. Keep in mind that the ideal person in this situation has undergone retraining. As shown in Fig4.(a), the ordinate is the optimal accuracy rate of each generation and the abscissa is the number of generations. It can be seen from the figure that in the 100th generation of the first stage, although the optimal accuracy fluctuates, it shows an overall trend of regional stability after an increase. After entering the second stage (100-150), the optimal accuracy rate began to continue to rise and finally stabilized. At the same time, we also tracked the evolutionary trajectory of the population. As shown in Fig4.(b), we divide the entire search process into five stages, that is, every 30 generations as a stage and check the distribution of the population in terms of accuracy and the number of parameters. It can be seen from the figure that the overall trend of the population is upward. In the early stage of evolution, the distribution of populations is relatively scattered. But at the end of the first stage, individuals are relatively concentrated and the upward trend of the population slows down. After entering the second stage, the distribution of individuals also becomes discrete and the accuracy rate of the population continues to rise. This shows that the two-stage strategy can help find better neural network architectures.

\begin{figure}
\centering
\setlength{\abovecaptionskip}{0.cm}
\includegraphics[width=85mm]{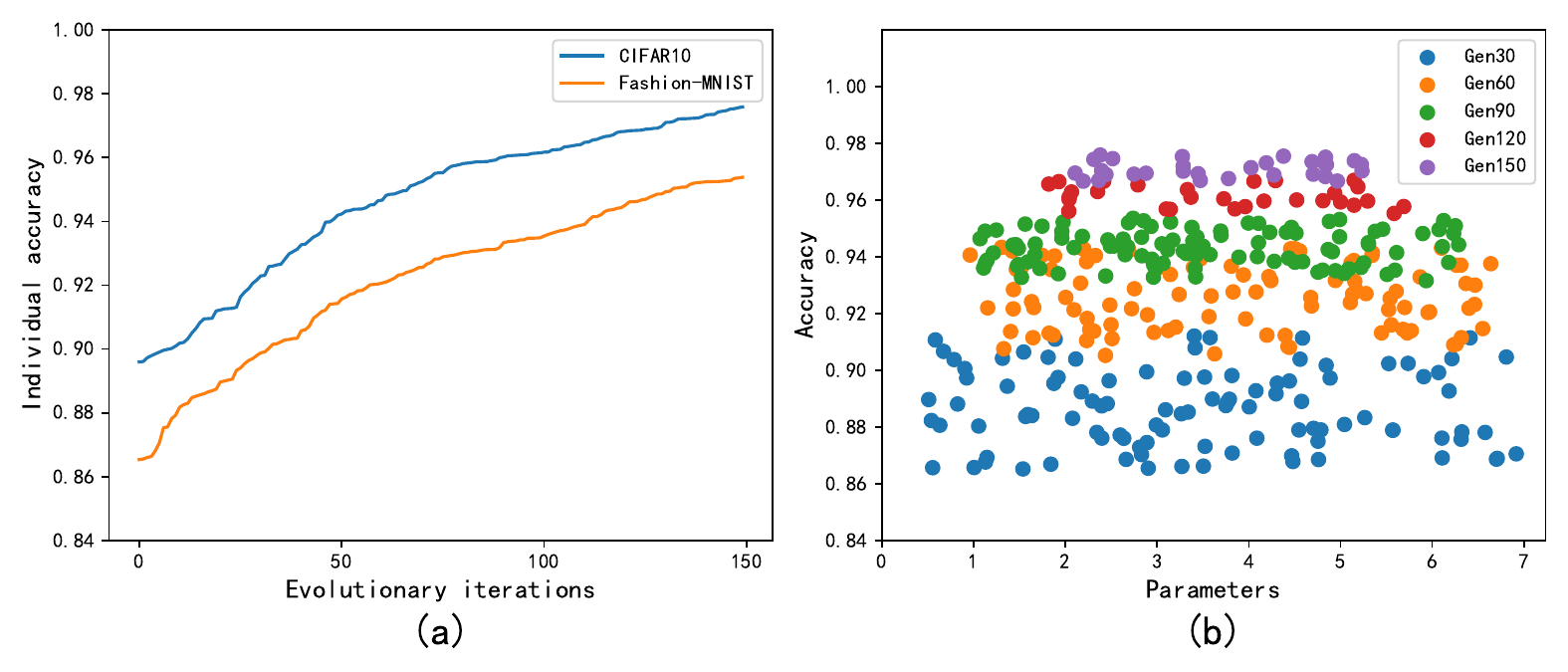}
\caption{(a) is the accuracy of the best individual in each generation is calculated on Fashion-MNIST and CIFAR10. (b) is population evolution trajectory diagram on CIFAR10.}
\label{fig1:env}
\end{figure}

\begin{figure}
\centering
\setlength{\abovecaptionskip}{0.cm}
\includegraphics[width=85mm]{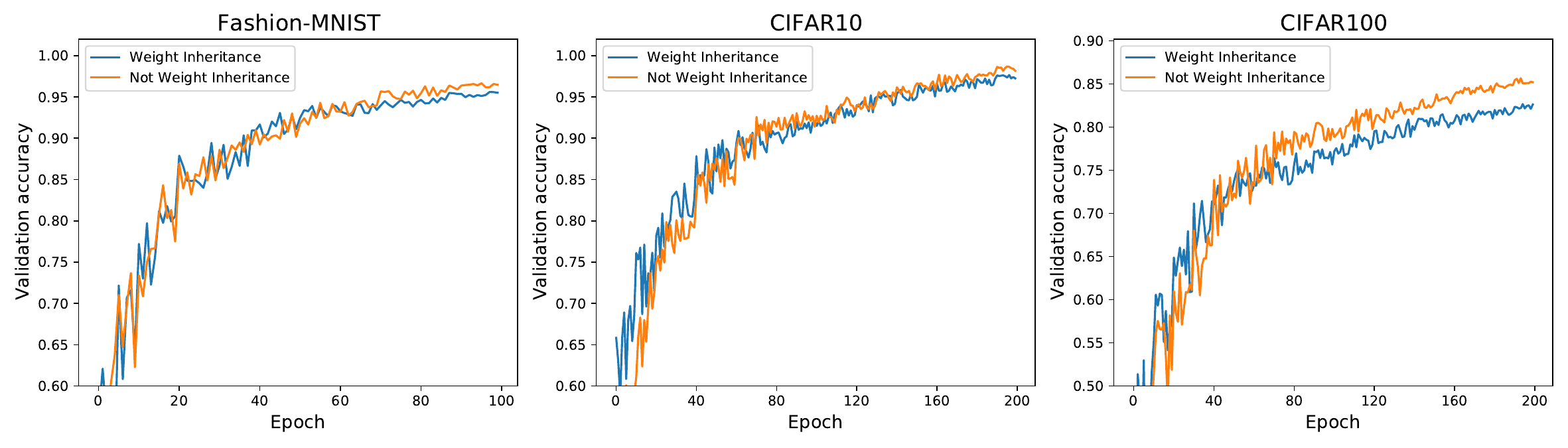}
\caption{The effectiveness of weight inheritance is compared on three benchmark datasets}
\label{fig1:env}
\end{figure}

\begin{figure}
\centering
\setlength{\abovecaptionskip}{0.cm}
\includegraphics[width=85mm]{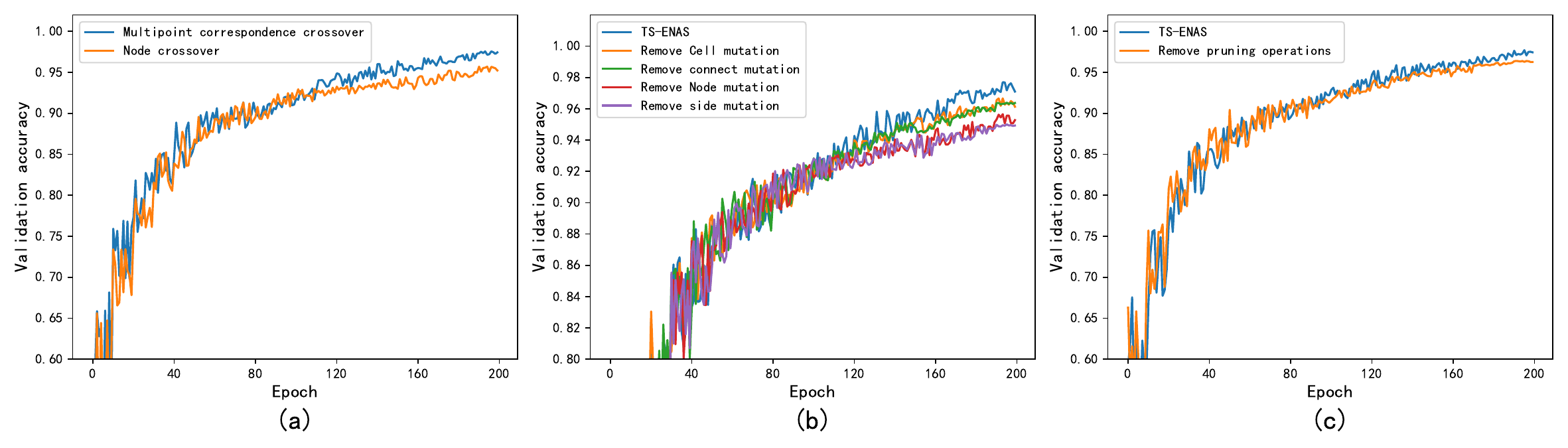}
\caption{The effectiveness of crossover, mutation and pruning operations is analyzed on the CIFAR10 dataset.}
\label{fig1:env}
\end{figure}

\subsection{Analysis of the effectiveness of the weight inheritance strategy}
Due to the high computational cost of NAS, many NAS algorithms employ acceleration methods. For ENAS, weight sharing is a good solution. TS-ENAS adopts the weight inheritance method based on cell to initialize the weight of the individual. For a new individual, only 10 epochs are trained for testing. Compared with the method of training each individual for 100 epochs, the efficiency is improved. Nearly 10 times. In order to verify the effectiveness of the weight inheritance strategy, we conducted comparative experiments. As shown in Fig5, we conducted comparative experiments on the three data sets of Fashion-MNIST, CIFAR10 and CIFAR100. Experimental results show that the optimal architecture found by the weight inheritance strategy is lossy in performance, but this loss is within an acceptable range.

\subsection{Effectiveness of crossover, mutation and pruning operations}

\subsubsection{Analysis of Crossover}
As previously introduced in Section IV-D, the crossover operation mainly performs the local search, which can help us better explore the search space. The crossover operation proposed in this paper is different from most existing ENAS algorithms. We use a multi-point corresponding crossover, that is, Normalcell only crosses with Normalcell and Reductioncell only crosses with Reductioncell. As shown in Fig6.(a), we validated the cross-operation on the CIFAR10 dataset. The results show that the crossover operation proposed in this paper can find a better neural network architecture than the single-point crossover.

\subsubsection{Analysis of Mutation}
The mutation is mainly to ensure the global exploration ability of the algorithm. In this paper, two mutation methods were used in the two stages of evolution. In order to verify the effectiveness of these mutation methods, we conducted a comparative experiment on the CIFAR10 dataset and the specific results are shown in Fig6.(b). The experimental results show that no matter which mutation operation is deleted, it will affect the final result. These four mutation operations can help find a better neural network architecture.

\subsubsection{Analysis of Pruning}
The individual is not transformed using the pruning process in the conventional EC approach. The number of internal nodes is fixed because the technique suggested in this study bases its initial stage on pre-defined cells. Exploring more diversified networks is made difficult by this. In order to decrease the number of nodes inside the cell, this study adds a pruning operation to the second stage of the search. We tested the trimming process on the CIFAR10 dataset in order to confirm its efficiency. The outcomes of the experiment are displayed in Fig.6(C). According to experimental findings, applying pruning operations can aid in the development of a more effective neural network architecture.

\section{CONCLUSION}
The TS-ENAS architecture focuses on the fixed number of network layers, the number of cells, and the connection mode of the cell-based NAS approach in order to discover the best neural network architecture within a broader search area. In a two-stage search strategy, the cells in the searched neural network structure are fine-tuned to find the best neural network structure, which expands the search space. First, an evolutionary algorithm is used to search for the general structure of the network based on the cell-based search space. The algorithm's searchability is guaranteed while its variety is also ensured. A novel cell-based search space and effective Double-coding are created to represent various building blocks and neural network architectures in the search space to match this two-stage search technique. We provide appropriate crossover, mutation, and pruning operators to further enhance the algorithm's searchability. A cell-based weight inheritance strategy is created to establish the network weights more quickly, considerably reducing the time required to evaluate the network architecture. On three benchmark classification problems from Fashion-MNIST, CIFAR10, and CIFAR100, the effectiveness of TS-ENAS is empirically confirmed. The performance of the technique is further tested on the ImageNet dataset in order to confirm its portability and effectiveness on large-scale datasets. According to experimental findings, the suggested TS-ENAS algorithm outperforms both the hand-designed networks and NAS algorithms already in use in terms of time and performance.

\newpage
 
\vfill

\end{document}